\documentclass{article}
\usepackage{spconf,amsmath,graphicx,hyperref}

\usepackage{algorithm}
\usepackage{algorithmic}
\usepackage{multirow}
\usepackage{booktabs}
\usepackage{amsmath}
\usepackage{amssymb}
\usepackage{tabularx}
\usepackage{makecell}

\title{MetaToolAgent: Towards Generalizable Tool Usage \\in LLMs through Meta-Learning}
\name{Zheng Fang$^{1,*}$, Wolfgang Mayer$^{2}$, Zeyu Zhang$^{1}$, Jian Wang$^{3}$, Hong-Yu Zhang$^{1}$, Wanli Li$^{1,\dagger}$, Zaiwen Feng$^{1,\dagger}$}

\address{
$^{1}$ College of Informatics, Huazhong Agricultural University, Wuhan, Hubei, China, 430070 \\
$^{2}$ Industrial AI Research Centre, University of South Australia, Adelaide, Australia, 5000\\
$^{3}$ School of Computer Science, Wuhan University, Wuhan, Hubei, China, 430072\\
}

\begin{document}
%
\maketitle
\begin{abstract}
Tool learning is increasingly important for large language models (LLMs) to effectively coordinate and utilize a diverse set of tools in order to solve complex real-world tasks. By selecting and integrating appropriate tools, LLMs extend their capabilities beyond pure language understanding to perform specialized functions. However, existing methods for tool selection often focus on limited tool sets and struggle to generalize to novel tools encountered in practical deployments. To address these challenges, we introduce a comprehensive dataset spanning 7 domains, containing 155 tools and 9,377 question-answer pairs, which simulates realistic integration scenarios. Additionally,  we propose MetaToolAgent (MTA), a meta-learning approach designed to improve cross-tool generalization. Experimental results show that MTA significantly outperforms baseline methods on unseen tools, demonstrating its promise for building flexible and scalable systems that require dynamic tool coordination. 

\end{abstract}

\section{Introduction}


The rapid emergence of large language model (LLM) agents has placed them at the forefront of research in the LLM domain, driving extensive investigation into their capabilities and applications. These agents go beyond conventional linguistic tasks by interacting with external tools\cite{lu2025toolfive}\cite{dong2025toolplaygrounds}, accessing APIs, and executing complex functions. Tooleyes\cite{ye2024tooleyes} suggest that the tool learning process in LLMs encompasses five dimensions: format alignment, intent comprehension, behavior planning, tool selection, and answer organization. Among these, tool selection is particularly crucial, as it requires the model to effectively retrieve and apply the optimal tool to satisfy the user’s request. Although current LLMs leveraging prompt engineering can handle most tool-related tasks successfully, real-world scenarios still occasionally result in incorrect tool choices.

To reconcile LLM with tool ecosystems, researchers have developed two distinct methodological paradigms: 

\textbf{Supervised Fine-Tuning (SFT)} uses annotated tool interaction datasets to train models for structured tool selection via direct parameter updates~\cite{ouyang2022training,schick2023toolformer}. While effective, it requires large amounts of high-quality labeled data, leading to high resource demands and often achieving limited performance in low-resource settings~\cite{zhou2024survey}. Toolformer~\cite{schick2023toolformer} addresses the lack of data by generating labels automatically but remains limited in tool diversity and computationally expensive.

\textbf{In-Context Learning (ICL)} avoids parameter updates by instead using few-shot examples and instructional prompts to implicitly guide models in tool invocation~\cite{ye2024tooleyes,shen2024hugginggpt,achiam2023gpt}. However, ICL’s non-parametric nature limits task-specific knowledge incorporation~\cite{PCRED}, and makes prompt optimization costly due to functional similarity among tools~\cite{ouyang2022training,zhou2024survey}.

Moreover, despite recent progress in tool-learning datasets, existing resources remain limited. In real-world vertical applications, we more often encounter two types of tool-invocation scenarios: (1) tools that need to be called across a wide range of scenarios for general-purpose LLMs, and (2) tools with subtle differences within the same domain, invoked by industry-specific LLMs. Most existing datasets do not categorize tools by usage scenario. For example, ToolAlpaca~\cite{ye2024tooleyes} covers over 400 tools but focuses on usage processes rather than tool selection. To address this gap, we compared several open-source datasets and constructed a dataset categorizing tools by usage scenario.

Motivated by these challenges---low-resource scenarios, fine-grained tool differentiation, and limited datasets---we propose \textbf{MetaToolAgent (MTA)}, a bi-level meta-learning framework~\cite{hospedales2021meta} that decomposes tool optimization into hierarchical objectives. Unlike traditional single-level optimization approaches based on fine-tuning or in-context learning, MTA solves an optimistic bi-level optimization (BLO) problem~\cite{LiuGZML22} where the outer-level generalization objective minimizes expected loss over unseen tasks: 

\begin{equation}
\min_{\phi} \mathbb{E}_{T_{\text{test}} \sim \mathcal{T}} \left[ \mathcal{L}_{T_{\text{test}}}(\theta^*(\phi), x, y) \right]
\end{equation}

with meta-parameters $\phi$ capturing cross-tool patterns across task distribution $\mathcal{T}$. The inner-level tool-specific objective refines policies for known tools: 

\begin{equation}
\theta^*(\phi) = \arg\min_{\theta} \mathcal{L}_{T}(\theta, x, y)
\end{equation}

The inner-level solutions $\theta^*(\phi)$ bound the outer generalization space while meta-gradients $\nabla_{\phi}$ refine $\phi$ through alternating updates ($\phi_{k+1} = \phi_k - \alpha \nabla_{\phi} \mathcal{L}_{T_{\text{test}}}$). By simulating dynamic tool ecosystems via this recursive BLO structure, MTA enhances semantic inference~\cite{jaiswal2020survey} and alignment capabilities, enabling robust orchestration without static prompts or task-specific fine-tuning through transfer optimization principles~\cite{tan2018survey}.

The main contributions of our work are summarized as follows:
\vspace{-0.2cm}
\begin{itemize}
    \item We construct a dataset comprising 155 tools across 7 scenarios and 9,377 user requests to evaluate LLMs' performance in tool selection.  
    \item We propose MTA, a meta-learning framework that enhances LLMs' ability to select appropriate tools when encountering previously unseen tools.  
    \item We evaluate multiple open-source LLMs of different scales and compare MTA with prompt engineering and fine-tuning approaches.  
\end{itemize}

\section{Method}


\subsection{Dataset Construction}


We construct our dataset via a systematic five-stage workflow, designed to ensure interpretability and effective training of tool-augmented LLMs:

\textbf{Scenario Definition:} We define seven real-world scenarios—office (Office), operating systems (OS), software development (SD), education (Edu), daily life (DL), Internet of Things (IoT), and mobile applications (App)—to guide tool creation and simulate diverse human-tool interactions.

\textbf{Tool Development:} We manually design 155 scenario-specific tools spanning the above domains, as summarized in Table~\ref{table3_2}.

\textbf{Tool Documentation:} Each tool is documented with its name, functionality, and input parameters, closely mirroring real-world API specifications.

\textbf{Query Generation:} Natural language queries that require tool calls are first generated by GPT-4 and then manually reviewed and refined to ensure clarity, correctness, and alignment with tool functionalities.

\textbf{Data Synthesis:} We synthesize structured data by parsing tool parameters from natural language queries, identifying candidate tools, and conducting tool selection analysis to determine the most appropriate tool for each query.

\begin{table}[ht!]
    \centering
    \small
    \caption{Statistics of Tool Quantity and Data Quantity}
    \begin{tabularx}{\linewidth}{c *{8}{>{\centering\arraybackslash}X}}
        \toprule
        & Office & OS & SD & Edu & DL & IoT & App & Total \\
        \midrule
        Tool & 20 & 21 & 24 & 26 & 20 & 22 & 22 & 155 \\
        Data & 1178 & 1288 & 1456 & 1548 & 1262 & 1276 & 1369 & 9377 \\
        \bottomrule
    \end{tabularx}
    \label{table3_2}
    \vspace{-0.3cm}
\end{table}

\subsection{MetaToolAgent Algorithm}

Our algorithm addresses the tool selection problem defined as follows: given a dataset $D = (Q, T)$, where $Q$ is a set of user queries reflecting various task requirements and $T$ is a collection of tools with their descriptions, the goal is to first sample a subset of tools $S \subseteq T$, and then select the correct tool $t \in S$ that matches a given query $q \in Q$.

To tackle this problem, our method fine-tunes the LLM using gradient descent. Unlike conventional gradient descent, which optimizes for a single objective per update, our approach structures each training iteration around a meta-task $M$, composed of multiple zero-shot tasks $\tau_i$. Each task $\tau_i$ is constructed using a fixed prompt template, together with the user query $q_i$ and descriptions of tools from $S$, the descriptions include definitions and usage information, and the task concludes with a fixed-format response that specifies the correct tool $t_i$. Model parameters are updated only after optimal performance is achieved on query $q_i$ within the meta-task framework, which distinguishes our approach from conventional fine-tuning methods.This design enables the model to learn generalizable features and strategies across varied tasks and tools.

\begin{algorithm}[t]
\caption{Fine-tuning LLM Based on Meta-Learning}
\label{algorithm1}
\renewcommand{\algorithmicrequire}{\textbf{Input:}}
\renewcommand{\algorithmicensure}{\textbf{Output:}}

\begin{algorithmic}[1] 
\REQUIRE Query dataset $\mathcal{Q}$, tool dataset $\mathcal{T}$, LLM parameters $\Theta$   
\ENSURE Updated model parameters $\Theta^*$     

\FOR{iteration = 1, 2, \ldots}
    \STATE Randomly sample a user query $q_i$ from query dataset $\mathcal{Q}$
    \STATE Find the correct tool $t_i$ corresponding to $q_i$ in the tool dataset $\mathcal{T}$.
    \FOR{k-step}
        \STATE From the tool dataset \( \mathcal{T} \), randomly sample a tool set \( S = \{ s_1, s_2, \ldots, s_j \} \) and insert the correct tool \( t_i \) to form the new tool set \( S = \{ s_1, s_2, \ldots, t_i, \ldots, s_j \} \).

        \STATE Construct the task \( \tau_k \) by integrating the user query \( q_i \) with the tool set \( S \)

    \ENDFOR
    \STATE Combine tasks $\{ \tau_1, \tau_2, \ldots, \tau_k \}$ to form a meta-task \( M \)
    
    \STATE Perform gradient descent on meta task \( M \)
    \STATE When optimal performance is achieved on user query \( q_i \), Update the model parameters as \( \Theta^i \leftarrow \Theta \).

\ENDFOR
\end{algorithmic}
\end{algorithm}

\subsection{Analysis of Generalization Capability in MetaToolAgent}

In this subsection, we analyze the generalization capability of MTA using the meta-learning framework of generalization error bounds\cite{shu2023learning}. The theoretical bound is formulated as:

\begin{equation}
\label{formulation_3_5}
R_(\hat{h}) - R_(h^*) \lesssim \tilde{\mathcal{O}}\left( \sqrt{\frac{C(\mathcal{H})}{\sum_{t=1}^T n_t}} + \frac{1}{T} \sum_{t=1}^T \sqrt{\frac{C(\mathcal{F})}{m_t}} + D_{\mathcal{F}} \right)
\end{equation}
\vspace{-0.3cm}

In the formula~\ref{formulation_3_5}, \( R(\hat{h}) \) represents the expected meta-generalization error of the learned meta-learner \( \hat{h} \), while \( R(h^*) \) denotes the expected error of the optimal meta-learner \( h^* \) within the hypothesis space \( \mathcal{H} \). The generalization error bound in meta-learning typically consists of three principal components. The first is the meta-learner complexity term, expressed as \( \sqrt{C(\mathcal{H}) / \sum_{t=1}^T n_t} \), where \( C(\mathcal{H}) \) denotes the complexity of the hypothesis space \( \mathcal{H} \) associated with the meta-learner, and \( n_t \) represents the number of query examples in the \( t \)-th training task. The second component is the task-specific learner complexity term, given by \( \frac{1}{T} \sum_{t=1}^T \sqrt{C(\mathcal{F}) / m_t} \), where \( C(\mathcal{F}) \) corresponds to the complexity of the hypothesis space \( \mathcal{F} \) for task-specific learners, and \( m_t \) denotes the number of support examples in the \( t \)-th task. The third component is the distribution discrepancy term, denoted \( D_{\mathcal{F}} \), which quantifies the distributional divergence between the support and query sets across tasks, evaluated under the hypothesis space \( \mathcal{F} \).

In MTA, we design different problem scenarios to optimize the query sets for each task, ensuring that each task has a sufficient number of query samples \( n_t \). This approach directly reduces the complexity term \( \sqrt{C(\mathcal{H}) / \sum_{t=1}^T n_t} \) of the meta-learner, helping to alleviate the impact of hypothesis space complexity, especially when the number of training tasks is limited. Furthermore, by introducing task-specific parameter adaptation mechanisms, we effectively leverage the rich prior knowledge embedded in pre-trained LLMs, thereby reducing the complexity \( C(\mathcal{F}) \) of task-specific learners. This, in turn, mitigates the risk of overfitting when the support set has a limited number of samples (i.e., when \( m_t \) is small). As a result, the task-specific learner complexity term \( \frac{1}{T} \sum_{t=1}^T \sqrt{C(\mathcal{F}) / m_t} \) is better managed, enhancing the robustness and generalization ability of task transfer.

\section{Experiments}

We conduct extensive experiments to verify the effectiveness of our framework and to answer the following research questions: \textbf{RQ1:} Is MTA universal across different models? \ \textbf{RQ2:} How does model size affect the baseline and MTA? \ \textbf{RQ3:} What are the advantages of MTA, and how does it differ from other methods?


\subsection{Experimental and Dataset Setup}
\label{Experimental Setup}

To evaluate the tool selection capabilities of MTA, we compare our approach with four baseline methods. \textbf{(1) Baseline (Base)} selects tools directly from prompts without additional reasoning guidance. \textbf{(2) Chain of Thought (CoT)}~\cite{wei2022chain} encourages the model to generate an explicit reasoning process before providing an answer. \textbf{(3) ReAct}~\cite{yao2022react} extends CoT by integrating reasoning with explicit instructions to query external information and incorporate feedback. \textbf{(4) Fine-tuning (FT)} employs the same prompt template as the Baseline but is trained via LoRA~\cite{hu2021lora}.

We evaluate four open-source LLMs covering diverse architectures and training backgrounds: ChatGLM3~\cite{glm2024chatglm}, Chinese-Alpaca-2~\cite{cui2023efficient}, Qwen1.5~\cite{yang2024qwen2}, and Baichuan2~\cite{baichuan2023baichuan2}.

To simulate practical tool selection scenarios, we use a proprietary tool dataset with two sampling strategies: single-domain (SD) and cross-domain (CD). The SD dataset contains tools exclusively from one domain, for example, if the test category is ``office'', all sampled tools belong strictly to that category. The CD dataset spans seven distinct domains, allowing tools from multiple categories to appear even when the target category is fixed—for instance, ``office'' tests may include tools from programming or mobile applications. In addition, we incorporate two public benchmark datasets, ToolAlpaca~\cite{toolalpaca} and API-Bench~\cite{API-Bench}, to evaluate the generalizability and robustness of our method across diverse tool-use scenarios.

During experiments, the output format for tool selection is explicitly defined. Model performance is evaluated based on both the correctness of the selected tools and adherence to the specified output structure. A response is considered correct only if it satisfies both criteria.

\subsection{Comparison Across Different LLMs}
\label{sec:Comparison Across Different Large Language Models}

For \textbf{RQ1}, we compare the tool selection capabilities of several different models using various methods. The results are in the table \ref{table4_4}.

Comparing sampling strategies, we find that SD demands more model reasoning capability because distinguishing similar tools is harder, which causes much lower performance than CD under the same settings. ICL methods like CoT and ReAct give only small improvements over baseline prompting, especially for weaker models such as Chinese-Alpaca-2-7B and Baichuan2-7B-Chat. In contrast, FT uses more resources but achieve better performance. Notably, MTA consistently improves tool selection capabilities across datasets, outperforming FT by a modest yet stable margin. On API-Bench, tool differences are subtle because scenarios are not clearly defined. Even Base or CoT achieves competitive results, and MFT gives further improvements. However, for ChatGLM3 and Baichuan2, MTA slightly underperforms FT because meta-task sampling increases the risk of overfitting when tool scenarios are not clearly distinct.

\begin{table}[h]
\centering
\caption{Scores of different methods evaluated on various models.}
\resizebox{\columnwidth}{!}{%
\begin{tabular}{p{1.5cm}cccccc}
\toprule
\textbf{Model} & \textbf{Dataset} & \textbf{Base} & \textbf{CoT} & \textbf{ReAct} & \textbf{FT} & \textbf{MTA} \\
\midrule
\multirow{4}{*}{\parbox{1.5cm}{ChatGLM3\\(6B)}}   
    & CD & 40.57 & 42.29 & 51.15 & 89.13 & \textbf{93.94} \\ 
    & SD & 35.35 & 36.27 & 40.68 & 78.68 & \textbf{83.68} \\
    & API-Bench & 50.23 & 55.12 & 65.34 & \textbf{83.12} & 82.45 \\
    & ToolAlpaca & 53.89 & 58.45 & 68.21 & 83.56 & \textbf{84.02} \\
\midrule
\multirow{4}{*}{\parbox{1.5cm}{Chinese-Alpaca-2\\(7B)} } 
    & CD & 29.86 & 26.41 & 25.82 & \textbf{92.57} & 91.46 \\ 
    & SD & 28.70 & 21.06 & 21.11 & 81.98 & \textbf{87.92} \\
    & API-Bench & 34.67 & 38.91 & 39.12 & 88.34 & \textbf{93.80} \\
    & ToolAlpaca & 36.23 & 40.88 & 44.39 & 90.12 & \textbf{91.03} \\
\midrule
\multirow{4}{*}{\parbox{1.5cm}{Qwen1.5-Chat (7B)}}     
    & CD & 40.82 & 44.34 & 57.55 & 96.39 & \textbf{97.16} \\ 
    & SD & 37.74 & 36.82 & 46.82 & 92.31 & \textbf{94.04} \\
    & API-Bench & 45.02 & 59.10 & 70.33 & 95.11 & \textbf{96.30}  \\
    & ToolAlpaca & 47.66 & 51.89 & 62.11 & 96.88 & \textbf{97.42} \\
\midrule
\multirow{4}{*}{\parbox{1.5cm}{Baichuan2-Chat (7B)}}   
    & CD & 39.17 & 34.48 & 46.20 & 96.89 & \textbf{97.33} \\ 
    & SD & 34.20 & 27.19 & 37.36 & 88.49 & \textbf{92.00} \\
    & API-Bench & 42.48 & 46.34 & 51.79 & \textbf{94.77} & 93.30 \\
    & ToolAlpaca & 44.23 & 48.45 & 53.45 & 95.12 & \textbf{95.70} \\
\bottomrule
\end{tabular}%
}
\label{table4_4}
\end{table}

\begin{figure}[ht]
\centering
\vspace{0.3cm}
\includegraphics[width=0.45\textwidth]{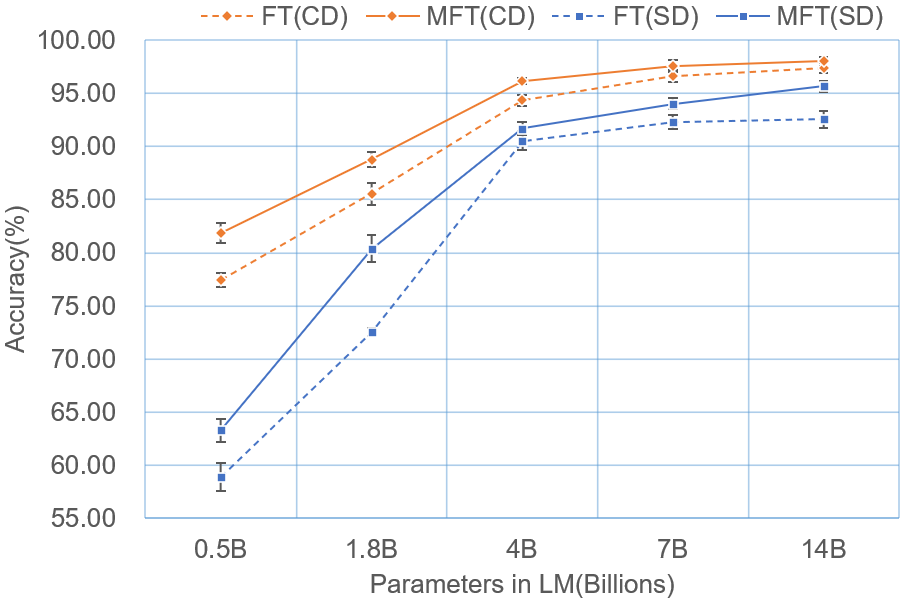} 
\caption{Accuracy of the Qwen model across different scales (0.5B, 1.8B, 4B, 7B, 14B) on CD (cross-domain) and SD (single-domain) datasets using MTA and FT (Fine-Tuning). }
\label{fig4_5}
\end{figure}



\subsection{Comparison on LLMs with Different Scales}
\label{sec:Comparison on Large Language Models of Different Scales}

For \textbf{RQ2}, to evaluate the applicability of our method across different model scales, we selected two models with varying sizes. Specifically, the Chinese-Alpaca-2 model offers only 7B and 13B parameter versions as open-source models, while Qwen1.5 provides 0.5B, 1.8B, 4B, 7B, and 14B parameter versions.

Figure \ref{fig4_5} shows that MTA consistently improves model performance over FT. For the Qwen series, the performance gap between MTA and FT narrows in the CD setting as model size increases, reflecting LLMs’ enhanced capacity to transfer knowledge across domains. In contrast, under the SD setting, the gap widens with larger model size, indicating that meta-learning becomes more effective at distinguishing between similar tools as model parameters grow.

\begin{figure}[ht]
\centering
\vspace{0.3cm}
\includegraphics[width=0.45\textwidth]{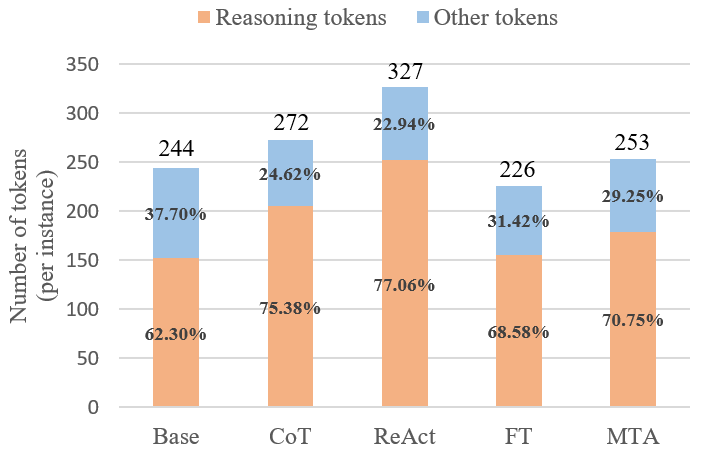} 
\caption{Average Token Distribution per Instance across Methods on ChatGLM3-6B}
\label{fig6}
\end{figure}

\subsection{Analysis of the Advantages and Limitations of Different Methods}
\label{sec:Case study}

For \textbf{RQ3}, we tested ChatGLM3-6B in a practical scenario and conducted an analysis of all the methods, with the results shown in Figure~\ref{fig6}.



CoT and ReAct tend to generate longer outputs with a higher proportion of reasoning tokens due to their step-by-step prompting design. In contrast, FT and MTA produce more concise responses, using fewer reasoning tokens while still maintaining task relevance. Qualitative analysis shows that the Base method often struggles to follow instructions or produce the required output format. While CoT and ReAct improve response structure, their verbosity reduces tool selection efficiency. FT generates cleaner, well-formatted answers with shorter prompts, demonstrating stronger alignment with task requirements. However, although FT is effective at identifying correct answers, it has difficulty distinguishing subtle differences between tools. MTA addresses this issue by enabling the model to more accurately recognize specific tool capabilities.

\section{Conclusion}




This work introduced MetaToolAgent (MTA), a meta-learning framework for LLM tool learning, and constructed a dataset spanning 7 scenarios and 155 tools. Through extensive experiments, we compared MTA against in-context learning and fine-tuning baselines. The results demonstrate that meta-learning offers substantial benefits. Specifically, MTA not only achieves higher accuracy but also exhibits stronger generalization to unseen tools, underscoring its capability to enhance adaptability and robustness of LLMs. These findings highlight the promise of meta-learning as a principled approach to advancing tool-use proficiency in LLMs, paving the way for more reliable and versatile deployment in real-world applications.

\vfill\pagebreak




\bibliographystyle{IEEEbib}
\bibliography{ICASSP}

\end{document}